\documentclass[letterpaper, 10 pt, conference]{ieeeconf}  

\IEEEoverridecommandlockouts                              

\usepackage{amsfonts}
\usepackage{graphicx}
\usepackage{biblatex}
\usepackage{float}
\usepackage{xcolor}
\usepackage[percent]{overpic} 
\usepackage{caption}
\usepackage{soul}

\title{\LARGE \bf
A Plenum Based Calibration Device for Tactile Sensor Arrays 
}

\author{Joan Kangro, Anand Vazhapilli Sureshbabu, Silvio Traversaro, Daniele Pucci and Francesco Nori
\thanks{*This work was supported by An.Dy Project which received funding from the EU Horizon 2020 research and innovation programme, under Grant 731540}
\thanks{All the authors are with Italian Institute of Technology, Italy
        {\tt\footnotesize kangrojoan@gmail.com, Anand.Vazhapilli@iit.it, silvio.traversaro@iit.it, Daniele.Pucci@iit.it, francesco.nori@iit.it}}%
}

\begin{document}

\maketitle
\thispagestyle{empty}
\pagestyle{empty}

\begin{abstract}

In modern robotic applications, tactile sensor arrays (i.e. artificial skins) are an emergent solution to determine the locations of contacts between a robot and an external agent. Localizing the point of contact is useful but determining the force applied on the skin provides many additional possibilities. This additional feature usually requires time consuming calibration procedures to relate the sensor readings to the applied forces. This paper presents a novel device that enables the calibration of tactile sensor arrays in a fast and simple way. The key idea is to design a plenum chamber where the skin is inserted, and then the calibration of the tactile sensors is achieved by relating the air pressure and the sensor readings. This general concept is tested experimentally to calibrate the skin of the iCub robot. The validation of the calibration device is achieved by placing the masses of known weight on the artificial skin and comparing the applied force against the one estimated by the sensors.

\end{abstract}

\section{INTRODUCTION} \label{intro_section}

Tactile sensor arrays, also known as artificial skins, are used in many fields of engineering including neuroprosthetics, humanoid robotics and wearable robotics \cite{lucarotti2013synthetic}. The artificial skins are usually mounted on the surface of robots in order to detect physical interactions with the external world. They enable the detection and localization of contacts using various sensor technologies such as capacitive, piezoresistive, piezoelectric, magnetic and optical \cite{girao2013tactile, yousef2011tactile}. 

Localizing the point of contact is useful but extracting the applied force intensity provides additional possibilities. For example, in the field of humanoid robotics knowing the contact forces could improve object manipulation (e.g. grasping), balancing, locomotion and human-robot interaction tasks\cite{cirillo2016conformable,parmiggiani2017design}. Using tactile skin to estimate the external forces also removes the need to use other, more expensive sensors (e.g. force-torque sensors).

In order to estimate the contact force, each tactile sensor within the array has to be calibrated to relate the sensor reading to the applied pressure, regardless of the operating principle of the sensor. In order to achieve that, a known pressure must be applied to each sensor to find a mathematical model between the applied pressure and the sensor reading. However, there are usually hundreds of sensors within a single tactile sensor array, making it a difficult and time consuming process.

This paper presents a novel device that enables the calibration of tactile sensor arrays so that the sensors could estimate the pressures applied to them individually. The calibration device is easy to set up, relatively fast (compared to other calibration techniques), and can be applied to a variety of skin shapes, sizes and technologies.

This paper is structured as follows. Section \ref{intro_section} introduces the topic of tactile sensor calibration and the device. Section \ref{background_section} describes the background of the work by giving a brief overview of the skin of iCub humanoid robot and presenting some related work. Section \ref{motivation_section} specifies the motivation behind the research. Section \ref{device_design} details the design of the calibration device, including conceptual, mechanical and electrical design. Section \ref{section_results} explains the obtained results and the validation procedure. Finally, Section \ref{discussion_section} concludes the paper by summing up the research and providing some guidelines for possible future work.

\section{BACKGROUND} \label{background_section}

\subsection{Skin of iCub} \label{icub_skin_description}

The experiments and validation of the proposed solution were done using the skin of the iCub robot. The iCub is a humanoid robot developed and manufactured in the Italian Institute of Technology. It is 104 cm tall, weighs around 22 kg and has 53 degrees of freedom (DOF). It has various sensors including inertial measurement units (IMU), force torque (FT) sensors, cameras, microphones, joint encoders and tactile sensor arrays, that cover the surface of the robot.

\begin{figure}[t!]
  \centering
  \vspace{0.4cm}
  \begin{minipage}[b]{0.4\textwidth}
    \includegraphics[width=\textwidth]{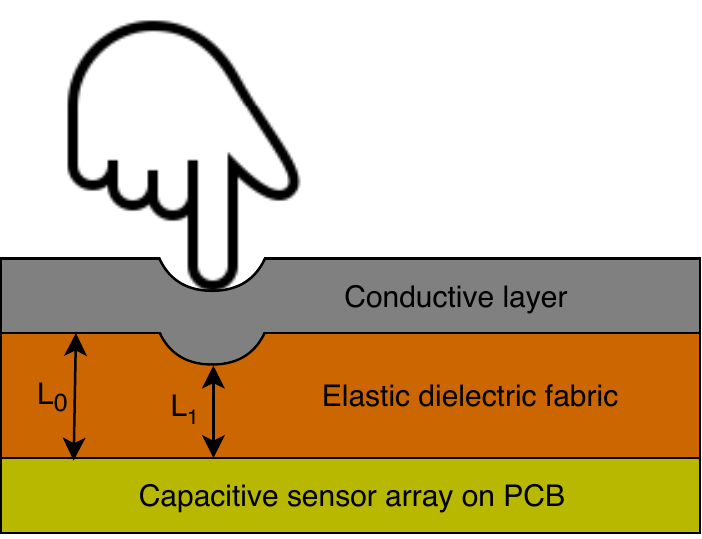}
    \caption*{(a) Main layers of the iCub skin.}
  \end{minipage}
  \hfill
  \begin{minipage}[b]{0.22\textwidth}
    \includegraphics[height=0.65\textwidth]{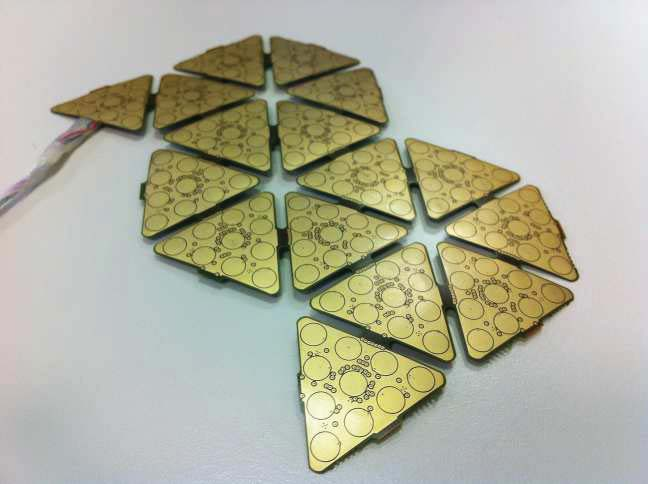}
    \caption*{(b) Triangular PCB modules that accommodate the capacitive sensors.}
    \end{minipage}
    \hfill
  \begin{minipage}[b]{0.22\textwidth}
    \includegraphics[height=0.65\textwidth]{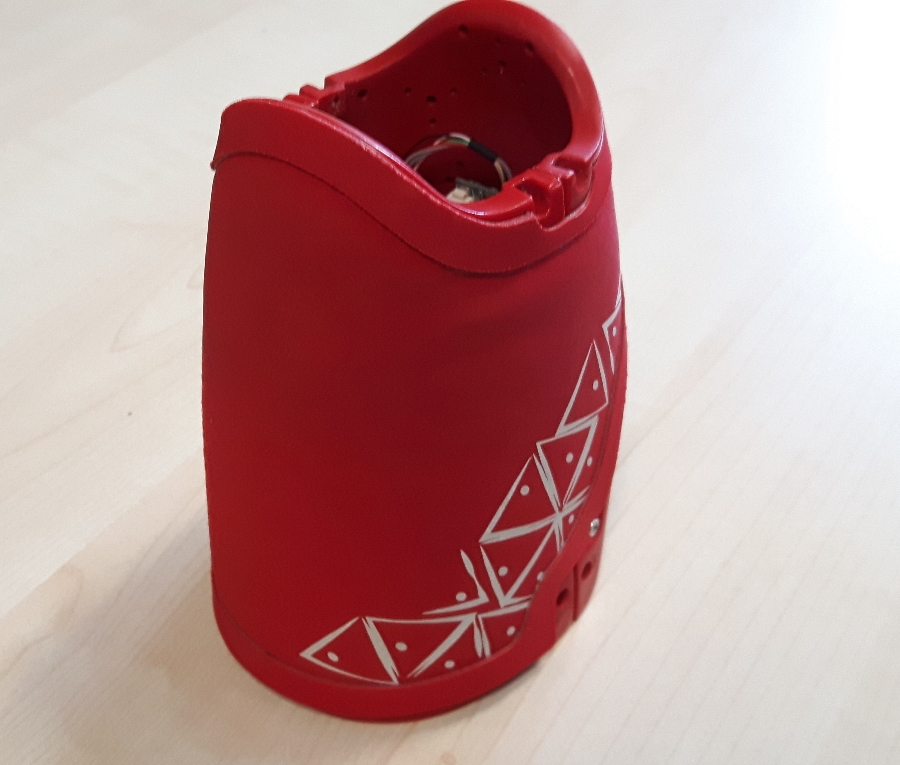}
    \caption*{(c) Assembled forearm skin of iCub.}
  \end{minipage}
  \caption{Tactile sensor array (i.e. skin) of iCub robot. It consists of 3 main layers: capacitive sensors on PCB (shown on (b)), elastic dielectric layer and the conductive layer. When the skin is touched, the distance between the capacitive sensors and the conductive layer decreases and therefore the capacitance changes that is measured by the sensors on the PCB (depicted on (a)). The assembled skin patch of left forearm of iCub is shown on (c).}
  \label{fig:skin_figures}
\end{figure}

The skin of iCub~\cite{Cannata2008} is an array of capacitive pressure sensors composed of flexible printed circuit boards (fPCB) covered by a layer of elastic fabric further enveloped by a thin conductive layer, as shown on Figure \ref{fig:skin_figures}(a). As the skin is touched (i.e. pressure is increased), the distance between the capacitive sensors and the conductive layer decreases and therefore the capacitance increases. However, the sensors output the inverted values of the capacitance, and therefore the raw capacitance values of the sensors tend to decrease as the pressure is increased. Each sensor has 8 bits of resolution.

The skin is composed of triangular modules of 10 sensors each (shown on Figure~\ref{fig:skin_figures} (b)), which act as capacitive pressure sensors, plus two temperature sensors for drift compensation. The tactile sensors have a measurable pressure range up to 180 kPa~\cite{bartolozzi2016robots}. The sensors' locations and orientations with respect to the robot frames are known.



The skin of iCub is divided into skin patches (also known as skin pieces) that consist of the aforementioned triangular modules. The iCub has skin patches for forearms, arms, hands, torso, upper and lower legs. For example, a skin patch of a left forearm of iCub is shown in Figure \ref{fig:skin_figures} (c). A single skin patch of iCub can have more than 500 individual tactile sensors. 

\subsection{Review of Literature} \label{literature_review_section}

There have been attempts to calibrate the robot skins and some of these methods are described in the following paragraphs.

One of the methods uses a technique that involves applying various forces mechanically on the individual tactile sensors with a device that enables the measurement of the applied forces \cite{o2015practical,maiolino2013flexible}. Therefore, it is possible to create the mathematical models that relate the applied force and the sensor values. However, all the methods that use this technique are very time-consuming, considering there can be hundreds of sensors within a single skin patch and each one of them has to be calibrated separately.

Another method to calibrate the robot's skin uses the force-torque (FT) sensors, located in the joints of the robot, that are able to measure the force and torque in all three dimensions. It uses the information from the skin and FT sensors together in order to approximate the stiffness of all the individual sensors \cite{ciliberto2014exploiting}. However, it requires the FT sensors to exist on the robots which is not always the case. 

Finally, a recent paper proposed a technique that applies uniformly distributed pressure on the skin to calibrate the skin \cite{kangro}. The skin was placed inside a vacuum bag and the pressure was decreased inside the bag with a vacuum pump. The pressure and skin values were extracted during the experiment and the models, that relate pressure to the sensor reading, were generated for all the sensors simultaneously. The calibration takes only a few minutes and can be applied to a variety of skin shapes.

\section{Motivation} \label{motivation_section}

Calibrating the tactile sensor arrays enables the measurement of contact forces that are applied to the robot. Using the tactile sensors to estimate external forces allows the accurate pinpointing of the location of the contact. 

The calibration is usually performed after the skin is installed on the surface of the robot as the installation induces variations in the performance of the sensors due to the curvature of the surface, variable thickness and stiffness of the dielectric layer, change in response due to the assembly discrepancies etc. In addition, the sensors deteriorate over time and should be re-calibrated after they exceed the acceptable error threshold. However, if desired, the proposed device can also be applied on the skin before installing it on the robot.

Most of the methods for calibration, described in Section \ref{literature_review_section}, are slow, as they require separate calibration for each sensor within the array. The skin patches of various robots consist of hundreds or even thousands of tactile sensors and therefore a calibration technique that is able to calibrate all the sensors simultaneously is desirable. 

In an attempt to calibrate all the sensors within an array simultaneously, introduced in Section \ref{literature_review_section}, a novel method was proposed that uses vacuum bags \cite{kangro} in which a pressure difference was induced on the opposite sides of the skin to apply a uniformly distributed pressure on the surface of the skin. A fifth-order polynomial model was used to relate the capacitance value to the applied pressure. 

It proved the concept but the experiment described had a number of drawbacks. Firstly, the setup introduced multiple problems such as air leaks and fluctuating pressure change during the calibration (as the pressure was regulated manually). Secondly, it pointed out that the pressure applied during the validation was often higher than the maximum pressure range for the calibration (using negative pressure for calibration constrains the maximum calibration range to be equal to atmospheric pressure, i.e. around 100 kPa) and therefore some sensors did not estimate the pressure correctly. It was also observed that the sensors saturated prematurely during the calibration due to the issues with air flow. Additionally, the experiment is relatively complicated to set up.

The device described in this paper allows the calibration of arrays of tactile sensors, that are used to estimate contact forces, by solving the issues mentioned in the previous paragraph. It enables the calibration procedure to be fast, accurate and achievable with a very simple setup. The pressure calibration range of this device is larger (300 kPa) and the sensors do not saturate prematurely, contrary to the vacuum bag experiment described above, and it can be used with skins of various shapes and sizes. The skin patch is simply inserted in the device and the connections from the skin are wired to a PC. The device then performs the calibration within a few minutes using the configurations desired by the user.

\section{DESIGN OF THE DEVICE} \label{device_design}

\subsection{Conceptual Design} \label{device_conceptual_design}

The calibration method involves using an isolating bladder to create a pressure difference on the opposite sides of the skin. This induces uniform pressure distribution applied on the surface of the skin. However, contrary to the vacuum bag experiment, explained in Section \ref{motivation_section}, where the pressure was decreased in the internal environment, the pressure is now increased in the external environment relative to the skin. Using positive pressure implies that the pressure range during calibration is not limited to atmospheric pressure as it is for the described vacuum bag experiment. 

\begin{figure}[t!]
\centering
\includegraphics[width=0.48\textwidth]{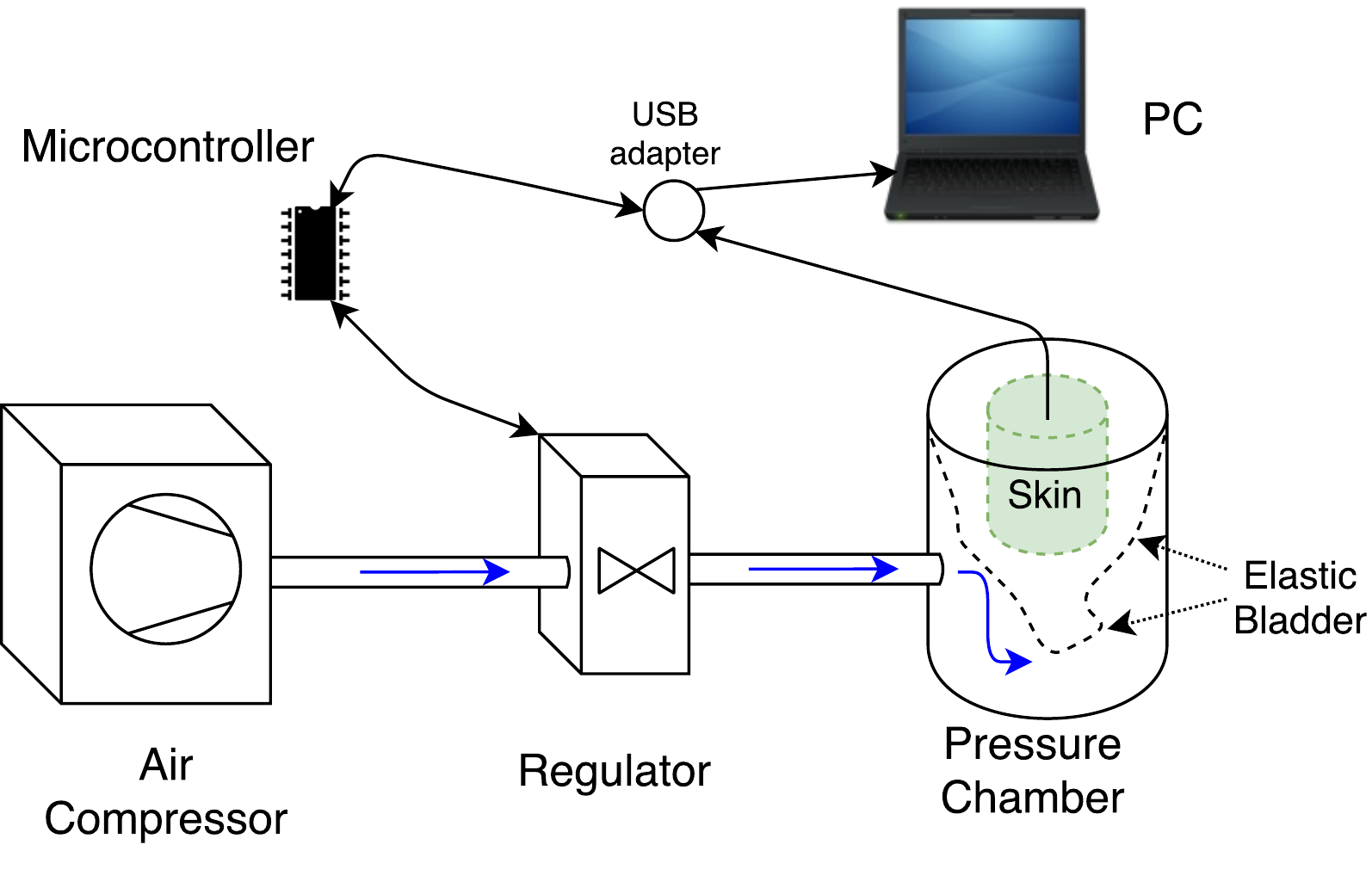}
\caption{Schematic of the conceptual design of the calibration device. The electrical connections are indicated with black solid arrows and the air flow with blue arrows. The air is pushed from the air compressor through a regulator to a pressure chamber. This forces the elastic bladder, inside the chamber, to wrap around the skin piece and induces uniformly distributed pressure on the skin piece. The pressure and sensor values are logged during the calibration to create a mathematical model that relates the sensor value and the pressure for each sensor.}
\label{fig:calibration_device}
\end{figure}

The sketch of the device is shown in Figure \ref{fig:calibration_device}. It consists of a microcontroller, a PC, an air compressor, a regulator, a pressure chamber and a compliant bladder. The compressor pushes air to the regulator that controls the pressure in its output. The regulator is able to increase the pressure with the rate that is required by the user until the desired maximum pressure is reached. The desired pressure is sent from PC through the microcontroller to the regulator and it separately measures the actual pressure on the output. As the air is pumped into the pressure chamber, the compliant bladder first wraps around the skin piece and then starts applying uniformly distributed pressure on the skin. The information from the skin about the tactile sensor values is also sent to the PC at the same time. 

The PC software gathers the skin sensory data and the pressure inside the chamber and logs it while the pressure is increasing. When the maximum desired calibration pressure is reached, the pressure is released in the chamber with the regulator. Then the gathered data is processed in order to create a mathematical model for each sensor that relates the applied pressure to the sensor capacitance value.

The mathematical model to relate the capacitance value to the applied pressure for each sensor is given by a fifth-order order polynomial as shown by 

\begin{equation} \label{eq:polynomial}
P(C_i) = a_{i} + b_{i} C_i + c_{i} C^2_i + d_{i} C^3_i + e_{i} C^4_i + f_{i} C^5_i
\end{equation}

where $P(C_i)$ is the pressure applied to a specific sensor and $a_{i},b_{i},c_{i},d_{i},e_{i},f_{i}$ are the sensor-specific constants representing the model for sensor $i$. The model is found by solving a least square optimization problem using the experimental data. This model was successfully tested during the vacuum bags experiment \cite{kangro}, described in Section \ref{motivation_section}. 

As the pressures applied on the individual sensors are known (using Equation \ref{eq:polynomial}), the total force applied on the skin patch can be calculated using a trilinear interpolation technique \cite{chavez2017contact} that includes the estimation of the pressures in between the sensors, given by

\begin{equation}
\vec{F}=\int_{v1}^{v2} \int_{u1}^{u2} p(u,v) \hat{{n}}(u,v) du dv, \label{eq:totalForce}
\end{equation} 

where $\vec{F}$ is the applied contact force, $p(u,v)$ is the pressure and $\hat{{n}}(u,v)$ is the skin's surface normal unit vector, the latter two defined over a coordinate system defined by u and v that represents the surface of the skin laid on a two dimensional plane. 

\subsection{Mechanical Design}

\begin{figure}[t!]
  \centering
  \vspace{0.4cm}
  \begin{minipage}[b]{0.48\textwidth}
  \centering
    \includegraphics[width=0.8\textwidth]{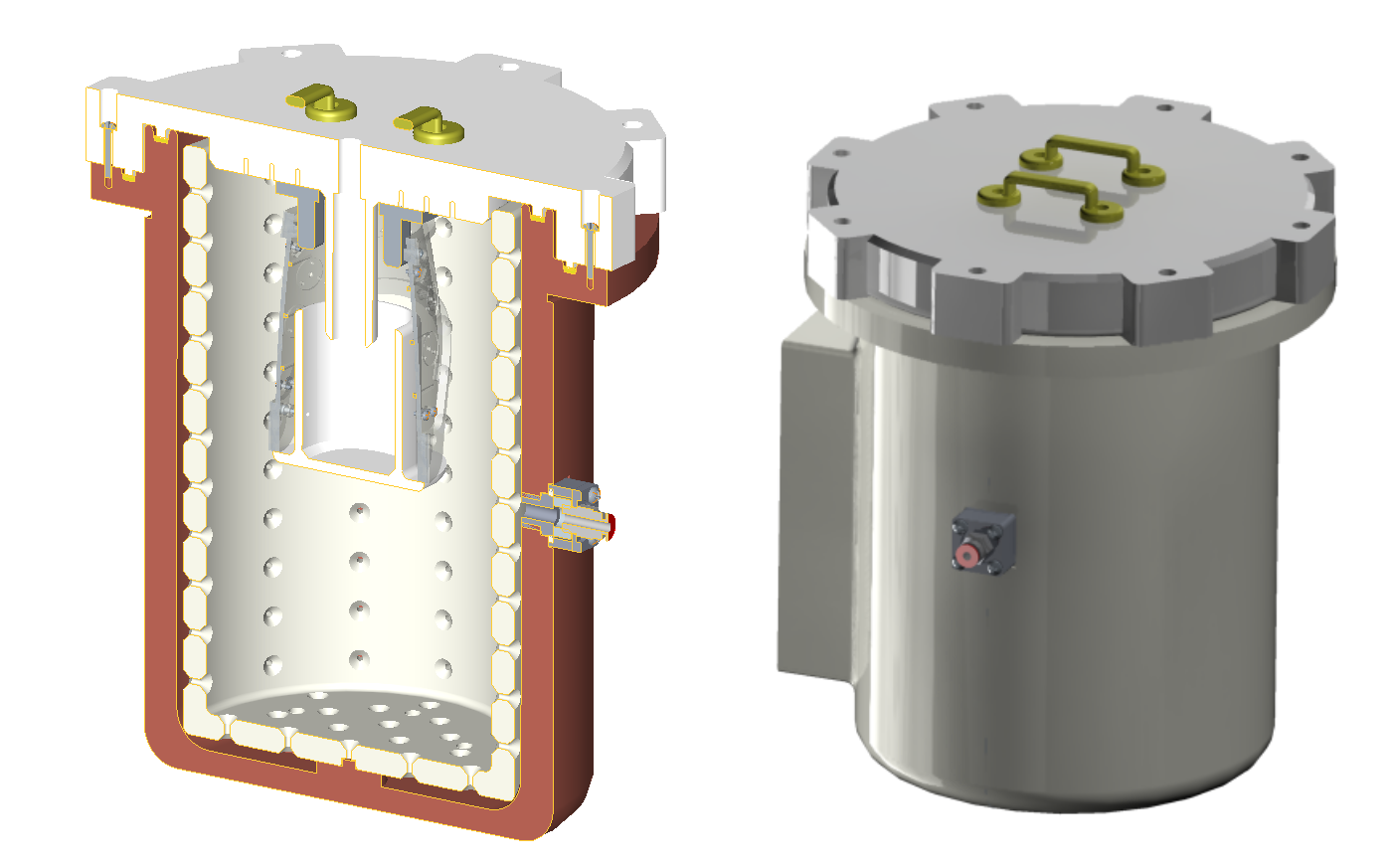}
    \caption*{(a) Cross-section and overall design of the device.}
  \end{minipage}
  \hfill
  \begin{minipage}[b]{0.48\textwidth}
    \includegraphics[width=\textwidth]{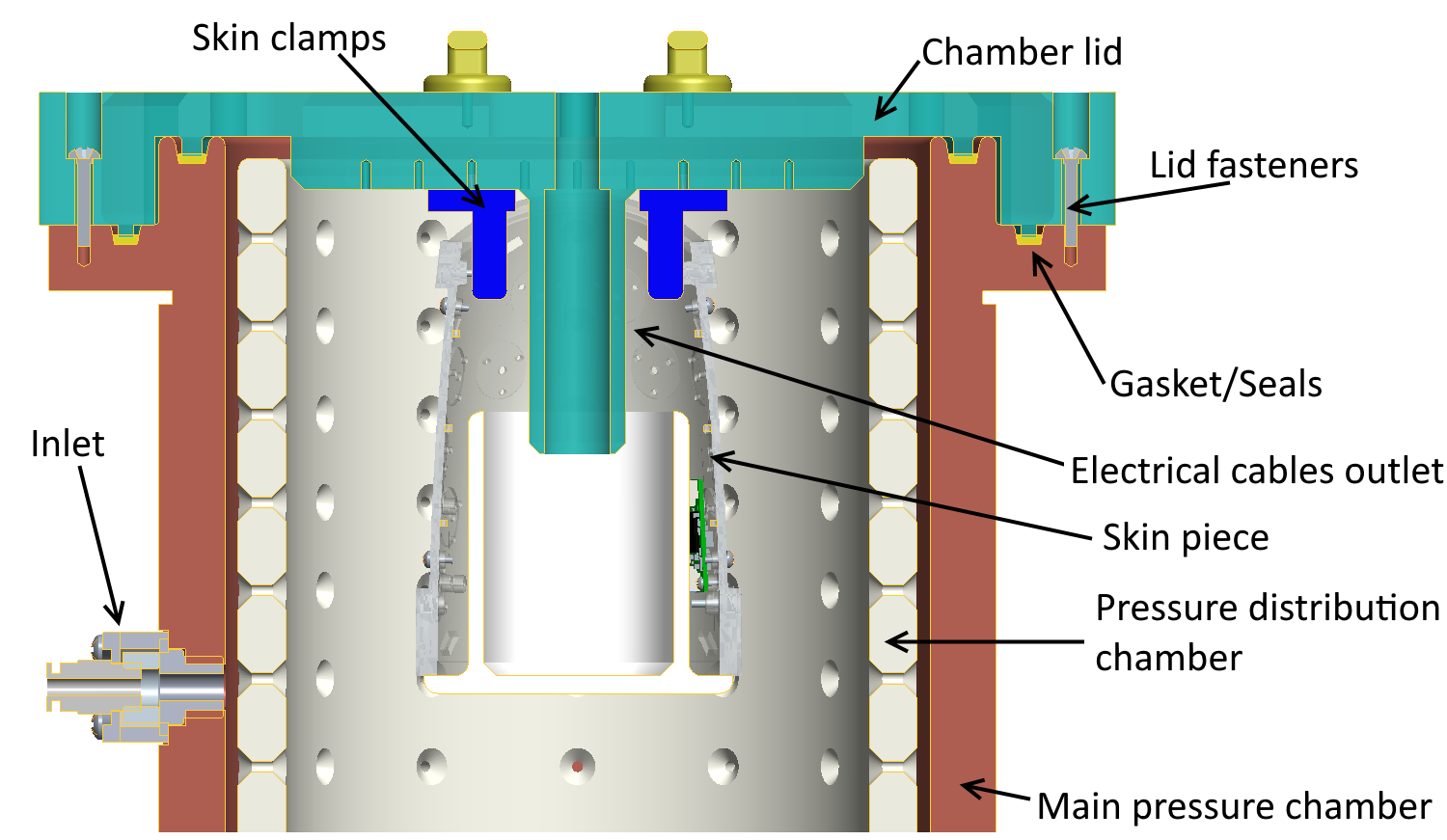}
    \caption*{(b) Detailed design with main components pointed out.}
  \end{minipage}
  \caption{Mechanical design of the calibration device.}
  \label{fig:mechanical design}
\end{figure}

The mechanical design is shown in Figure \ref{fig:mechanical design}. It includes a pressure chamber, a pressure distribution component, a lid, an inlet nozzle and a bladder. The device is manufactured using ABS material with fused deposition modeling (FDM), an additive manufacturing method. After manufacturing the device, it was coated on the inside using a food-grade silicone to prevent any kind of leaks that could arise due to additive manufacturing.

The pressure chamber is a cylindrical and hollow component that has to be able to withstand pressures up to 300 kPa. The pressure distribution component is mounted in the chamber in order to distribute the pressure in the chamber uniformly. However, it was observed during the experiments that in static conditions the pressure distribution component did not change the performance of the device.

The skin component that is being calibrated is mounted onto the lid of the pressure chamber. The lid contains a hollow shaft to route the electrical wires from the sensor boards of the skin piece. Using the dedicated holes on the skin component, it is attached onto the lid using custom prototyped mounts. The lid also has projections which mate with the recesses provided in the main chamber.

To minimize the air leaks between the lid and the main chamber, the device is provided with a two step hermetic seal. The gasket was rapid-prototyped using a rubber-like material called TANGO \cite{tango-ref} that goes in the recesses of the pressure chamber. This means the device is securely sealed by fastening the lid to the main chamber. 

In order to induce a pressure difference between the opposite sides of the skin piece, a deformable bladder was used. It is placed inside the chamber and fixed by attaching the lid to the chamber. A picture of the bladder inside a chamber is shown of Figure \ref{fig:experiment} (a). When the air is pushed in from the inlet nozzle, the bladder first wraps around the skin component and then starts applying uniformly distributed pressure on the skin piece.

\subsection{Electrical Design}

\begin{figure}[t!]
\centering
\includegraphics[scale=0.48]{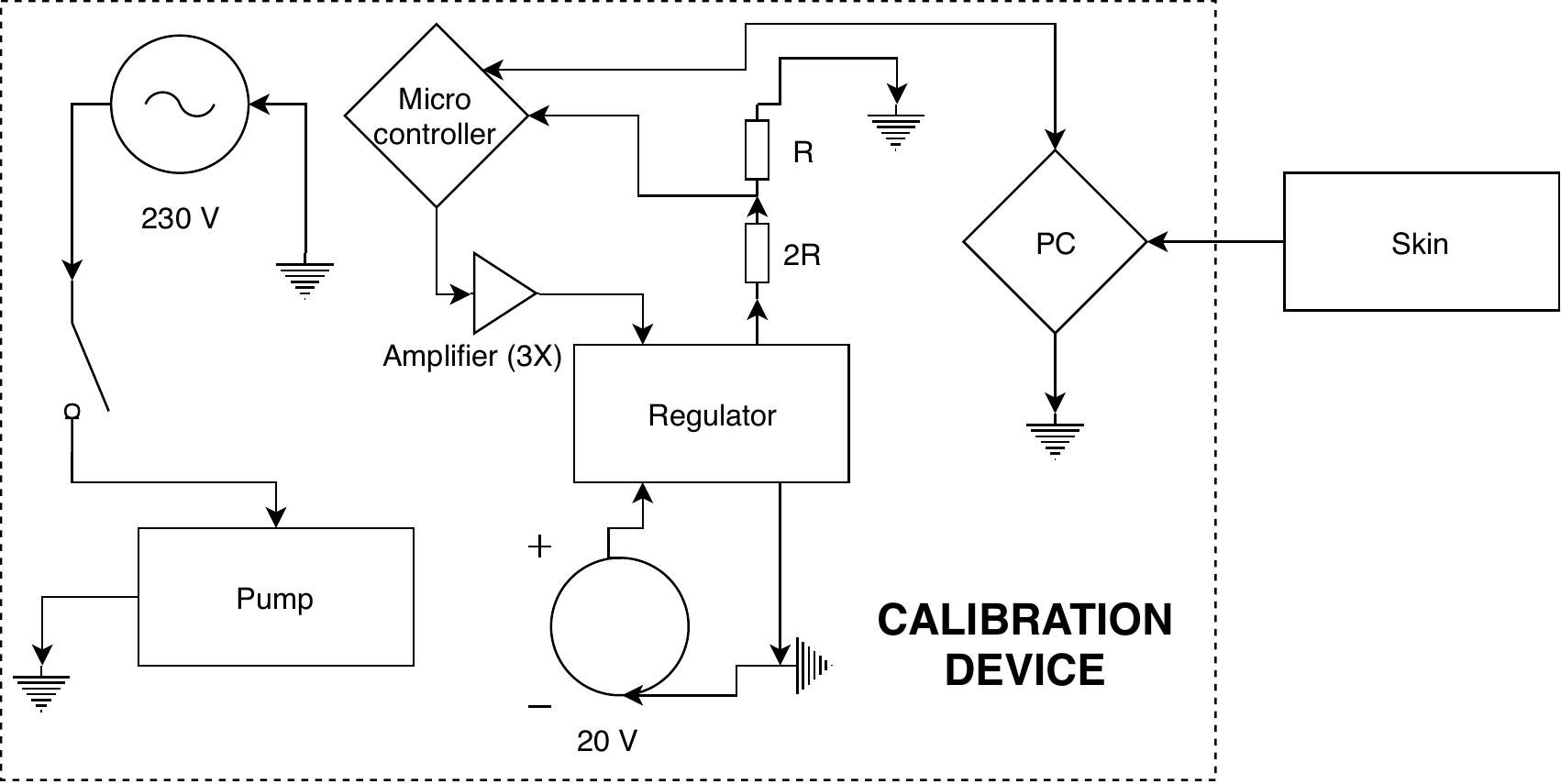}
\caption{Schematic of the electrical design of the calibration device.}
\label{fig:electronics}
\end{figure}

The electrical schematic of the device is presented in Figure \ref{fig:electronics}. A microcontroller is connected to the regulator using two separate pins. One of those pins controls the desired pressure using an analogue signal through a non-inverting amplifier (ratio of 3) because the analogue range of the microcontroller and the regulator are 3.3 volts and 10 volts respectively. The other pin indicates the actual pressure in the output of the regulator using an analogue signal that is scaled down 3 times using a simple voltage divider circuit using high value resistors. The microcontroller is also connected to a PC, using a serial communication protocol, where the main software program is running. The regulator is powered by 20 volts DC supply and the compressor uses 230 volts AC supply. In order to use the device, a skin is simply connected to the PC in order to send the tactile sensor values (currently using the CAN communication protocol). 

\section{RESULTS} \label{section_results}

\begin{figure}[t!]
    \centering
    \vspace{0.4cm}
    \begin{minipage}[b]{0.21\textwidth}
        \includegraphics[height=1.2\textwidth]{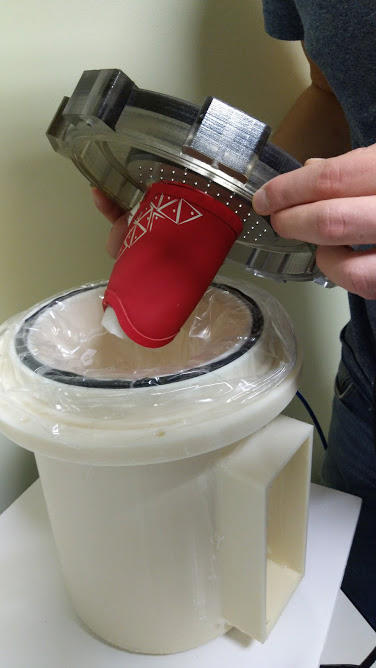}
        \caption*{(a) Skin attached to the lid is placed inside the chamber.}
    \end{minipage}
    \hfill
    \begin{minipage}[b]{0.265\textwidth}
        \includegraphics[width=\textwidth]{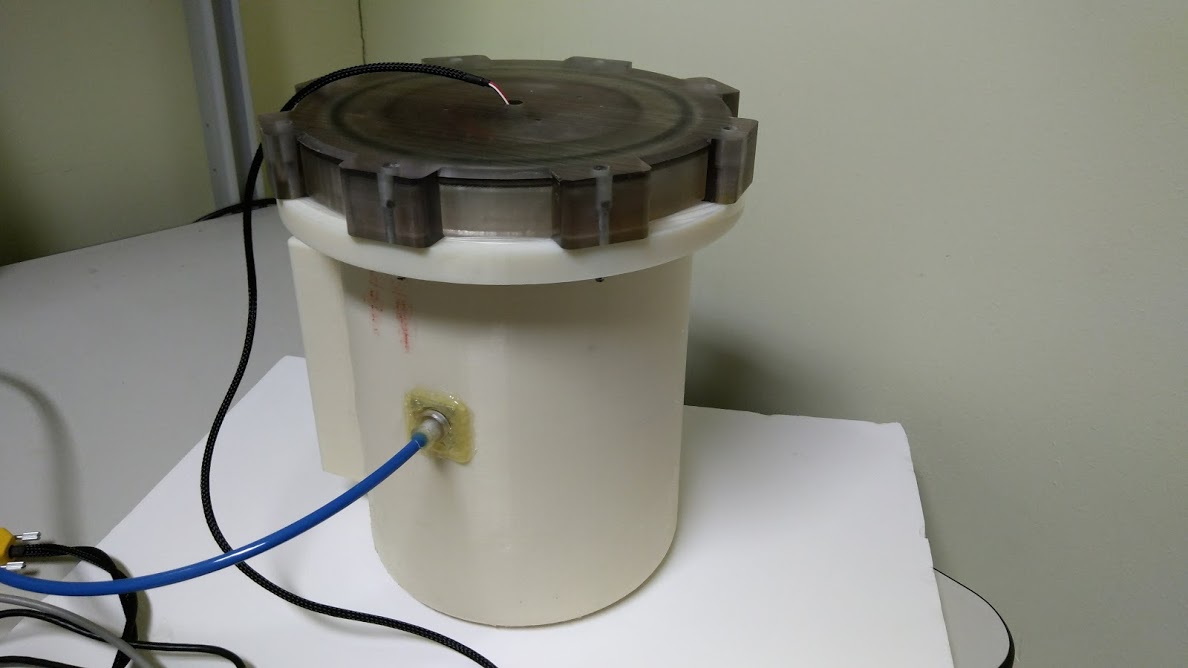}
        \caption*{(b) Pressure chamber. The air is pushed to the chamber through the small tube and the data from the skin is extracted using the wire coming out from the top.}
    \end{minipage}
    \hfill
    \caption{Pictures of the implementation of the device.} 
    \label{fig:experiment}
\end{figure}

\begin{figure*}
	\includegraphics[width=\textwidth]{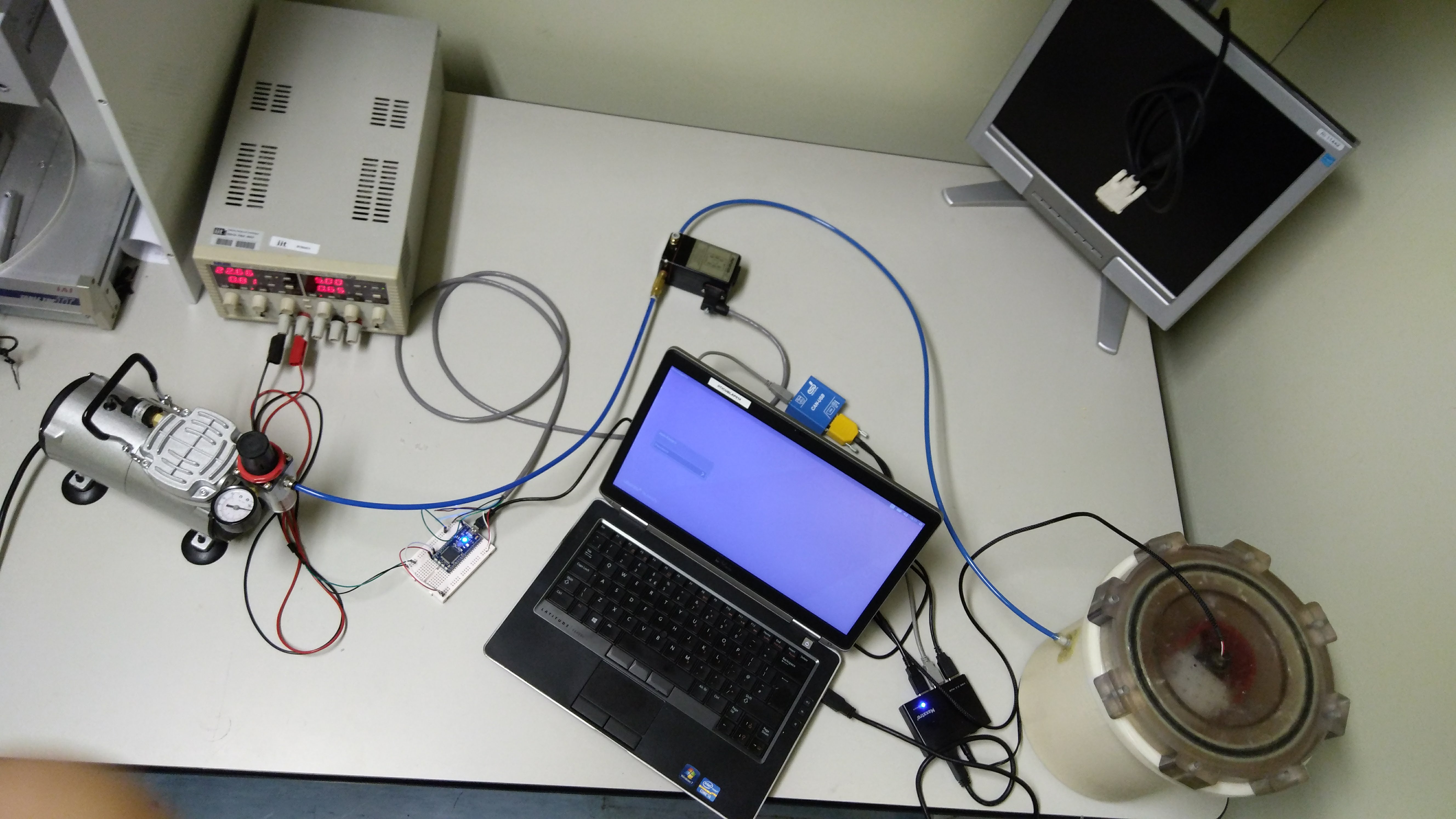}
	\put(-493,60){\color{black}Compressor}
    \put(-468,72){\color{black}\vector(0,1){27}}
    \put(-325,225){\color{black}Regulator}
    \put(-290,220){\color{black}\vector(1,-1){15}}
    \put(-40,95){\color{black}Chamber}
    \put(-20,90){\color{black}\vector(-1,-1){20}}
    \put(-380,30){\color{black}Microcontroller}
    \put(-350,40){\color{black}\vector(0,1){28}}
    \put(-145,140){\color{black}PC}
    \put(-148,137){\color{black}\vector(-1,-1){20}}
	\caption{Full calibration setup with the main components pointed out.}
	\label{fig:experiment_setup}
\end{figure*}

The device was implemented as shown on Figure \ref{fig:experiment} and \ref{fig:experiment_setup}. The forearm skin of iCub was placed in the device and the results of the calibration are described in the sections below. 

\subsection{Calibration Results}

During the calibration, the sensor and the pressure values were simultaneously recorded at 10 Hz. The data was extracted while the pressure was increasing and stopped when the maximum pressure was reached in order to avoid the hysteresis effect \cite{kangro}. The hysteresis effect occurs when the pressure is decreasing because it takes time for the flexible fabric to return to its actual position for a given pressure. 

\begin{figure}[t!]
    \centering
    \includegraphics[height = 0.6\linewidth]{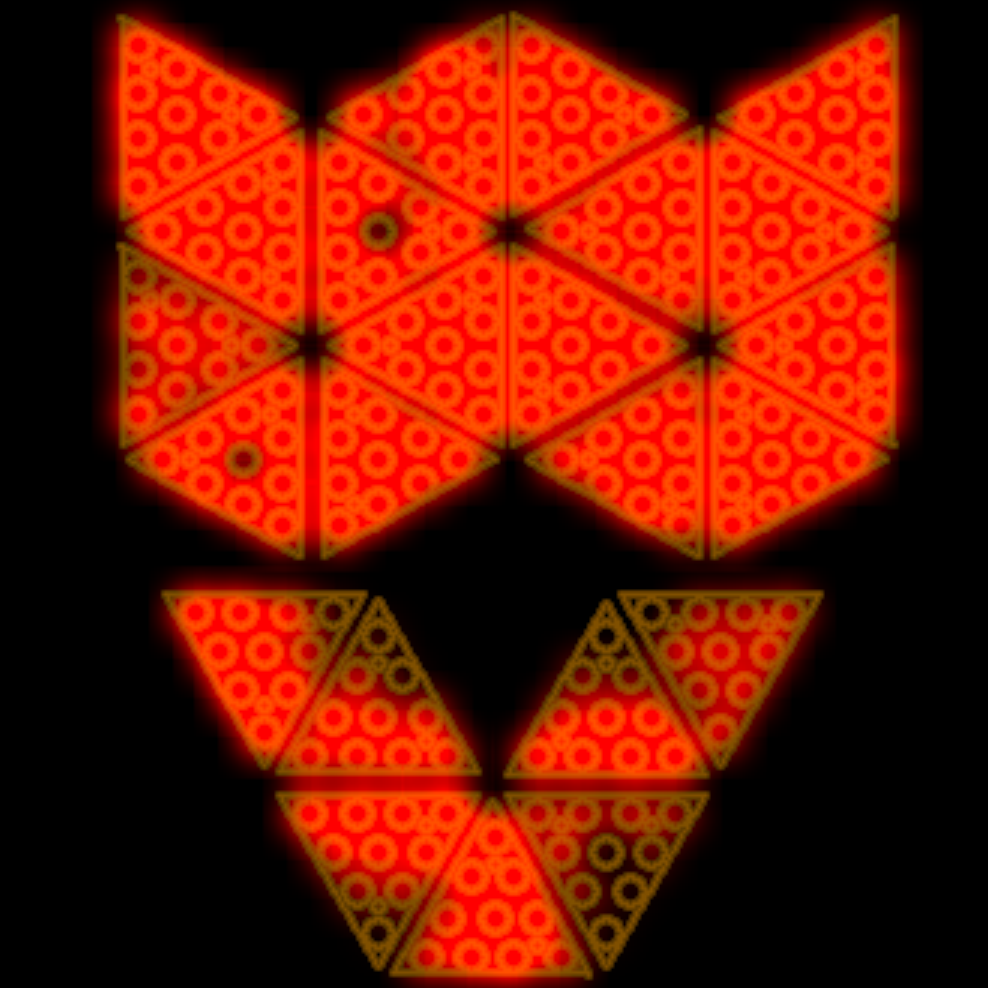}
    \caption{Map of skin response while the skin is being calibrated. The brightness of the red color indicates the change in capacitance value. It can be seen that the sensors have different responses to the uniform applied pressure.}
    \label{fig:skin-response}
\end{figure}

Figure \ref{fig:skin-response} shows how the capacitance values have changed when the maximum pressure is applied. It can be observed that the capacitance has changed for most of the sensors. The sensors on the bottom of the figure that have not significantly changed are the ones that are cut (required during the assembly process) and therefore have much lower sensitivity. The observation that the brightness of the sensors is not uniform indicates that the capacitance change is not the same among the sensors. This means that sensors have different responses to the same applied pressure, which corroborates that each sensor should have its own mathematical model to relate the applied pressure and the sensor reading.

\begin{figure}[t!]
  \centering
  \vspace{0.4cm}
  \begin{minipage}[b]{0.45\textwidth}
    \includegraphics[width=\textwidth]{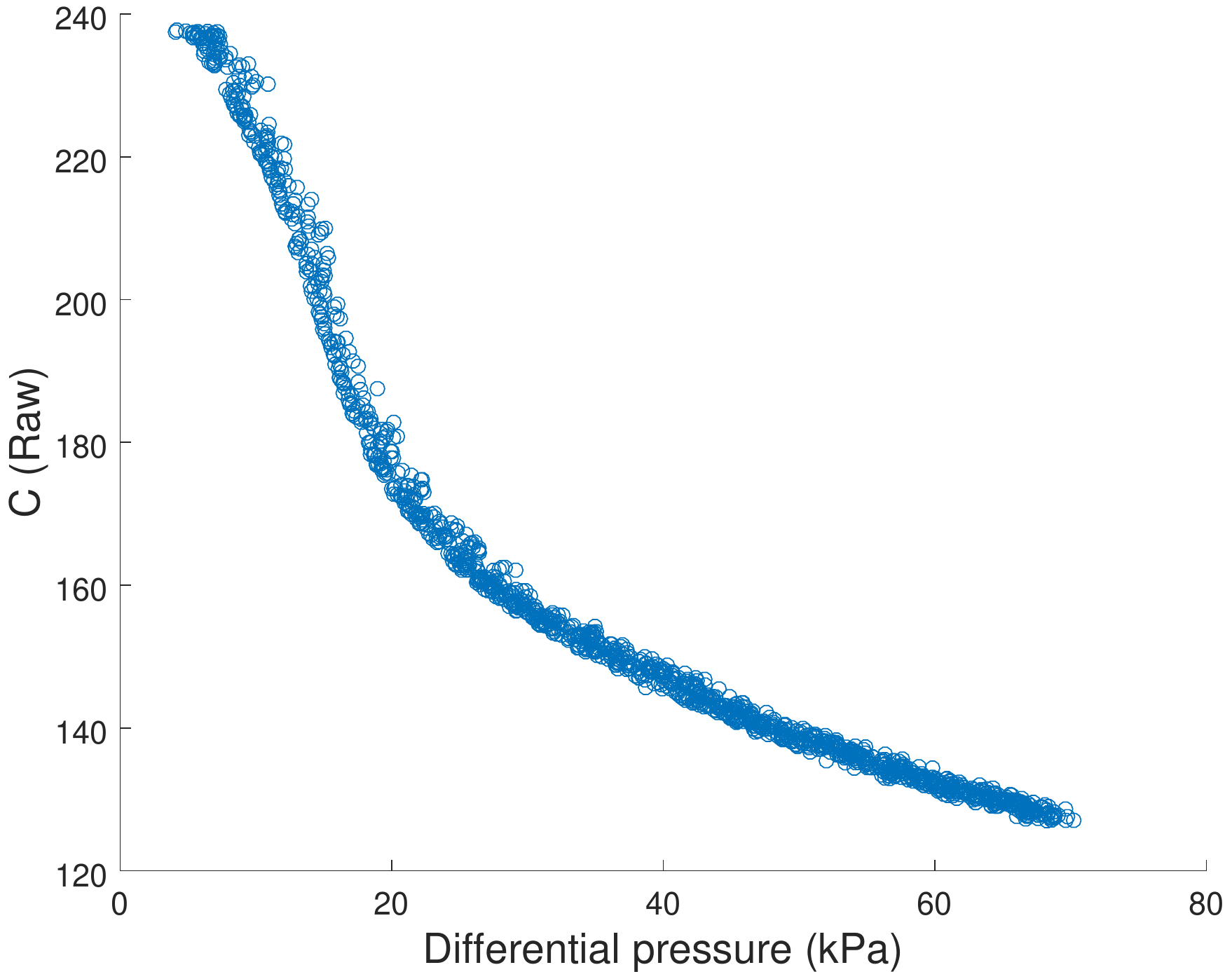}
    \caption*{(a) using the calibration device.}
  \end{minipage}
  \hfill
  \begin{minipage}[b]{0.45\textwidth}
    \includegraphics[width=\textwidth]{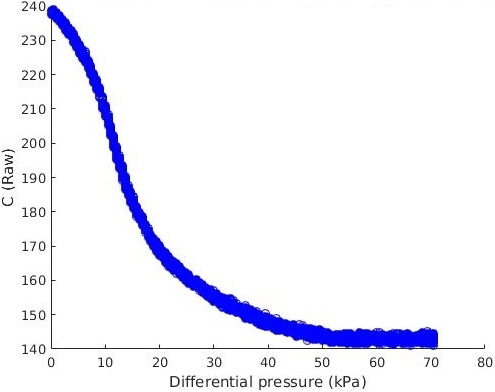}
    \caption*{(b) using the vacuum bags setup.}
  \end{minipage}
  \caption{Average sensors' capacitance response versus pressure. The comparison between the calibration device and vacuum bags experiment indicates similar behaviour by the sensors. However, the sensors saturated at around 55 kPa during the vacuum bag experiment.}
  \label{fig:calibration}
\end{figure}

To evaluate the performance of the device, the gathered calibration data was compared to the results from the vacuum bags experiment, described in Section \ref{motivation_section}. The same skin piece was used in order to keep all the other specifications for the problem the same. As seen from Figure \ref{fig:calibration}, the curves, indicating the average sensor response as the pressure increases for both calibration procedures, have a very similar shape. However, during the vacuum bags experiment the sensors saturated at around 55 kPa but this was not observed during the calibration with the device. The saturation during the vacuum bags experiment might have been caused due to air flow problems that could have occurred because of the experimental setup.

\begin{figure}[t!]
\centering
\includegraphics[width=0.45\textwidth]{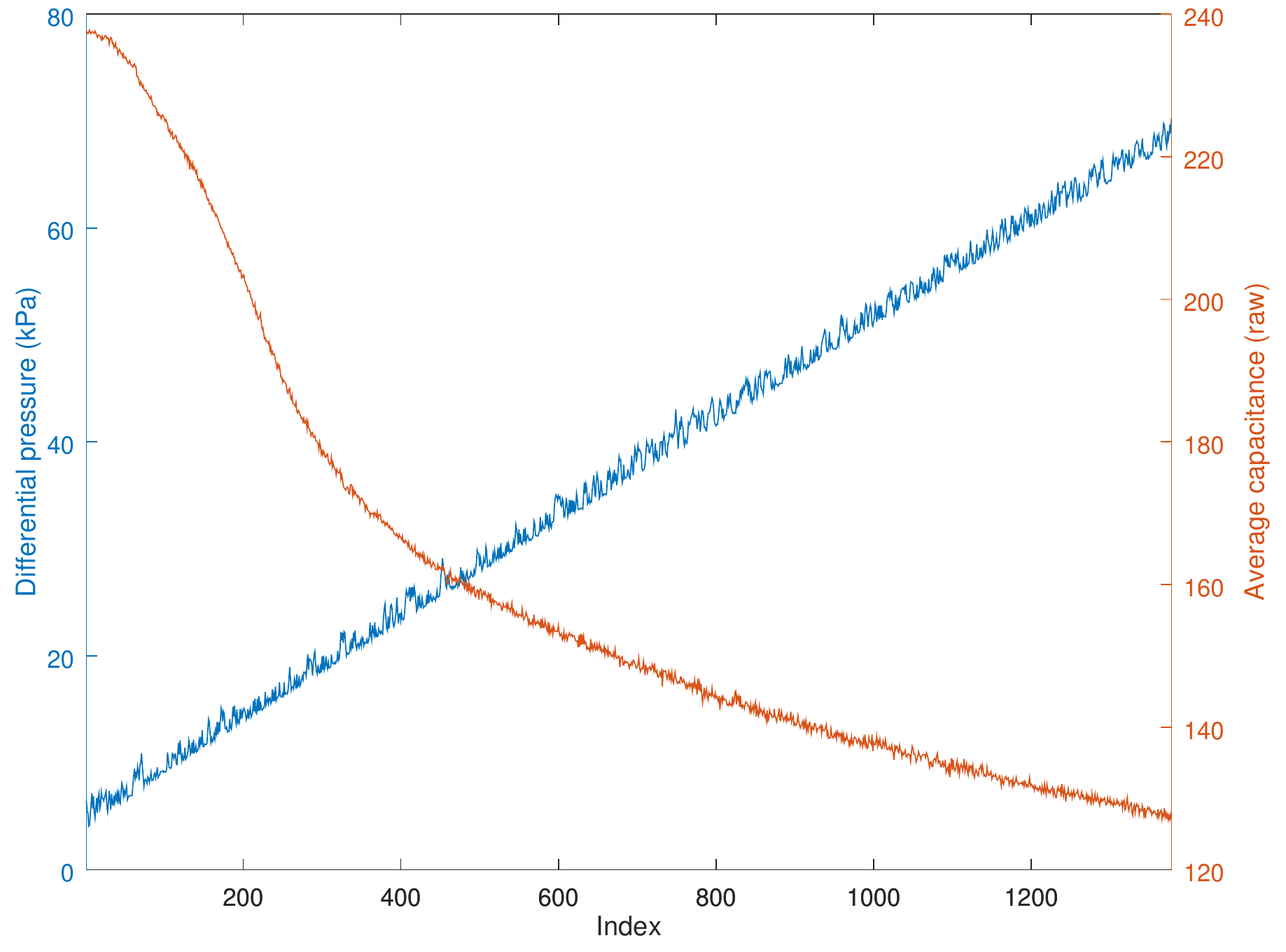}
\caption{Pressure and average capacitance during the experiment. The pressure is increased with a constant rate during the calibration and the sensors exhibit inversely proportional relationship.}
\label{fig:pressure_and_capacitances}
\end{figure}

Figure \ref{fig:pressure_and_capacitances} shows how the pressure and average raw capacitance measurements change during the calibration with the device. The plot indicates that the raw capacitance and pressure a have roughly inversely proportional relationship. The pressure value is increased incrementally throughout the experiment until the maximum pressure is reached. However, it can be seen that the pressure sensor reading is relatively noisy.

\begin{figure}[t!]
\centering
\includegraphics[width=0.45\textwidth]{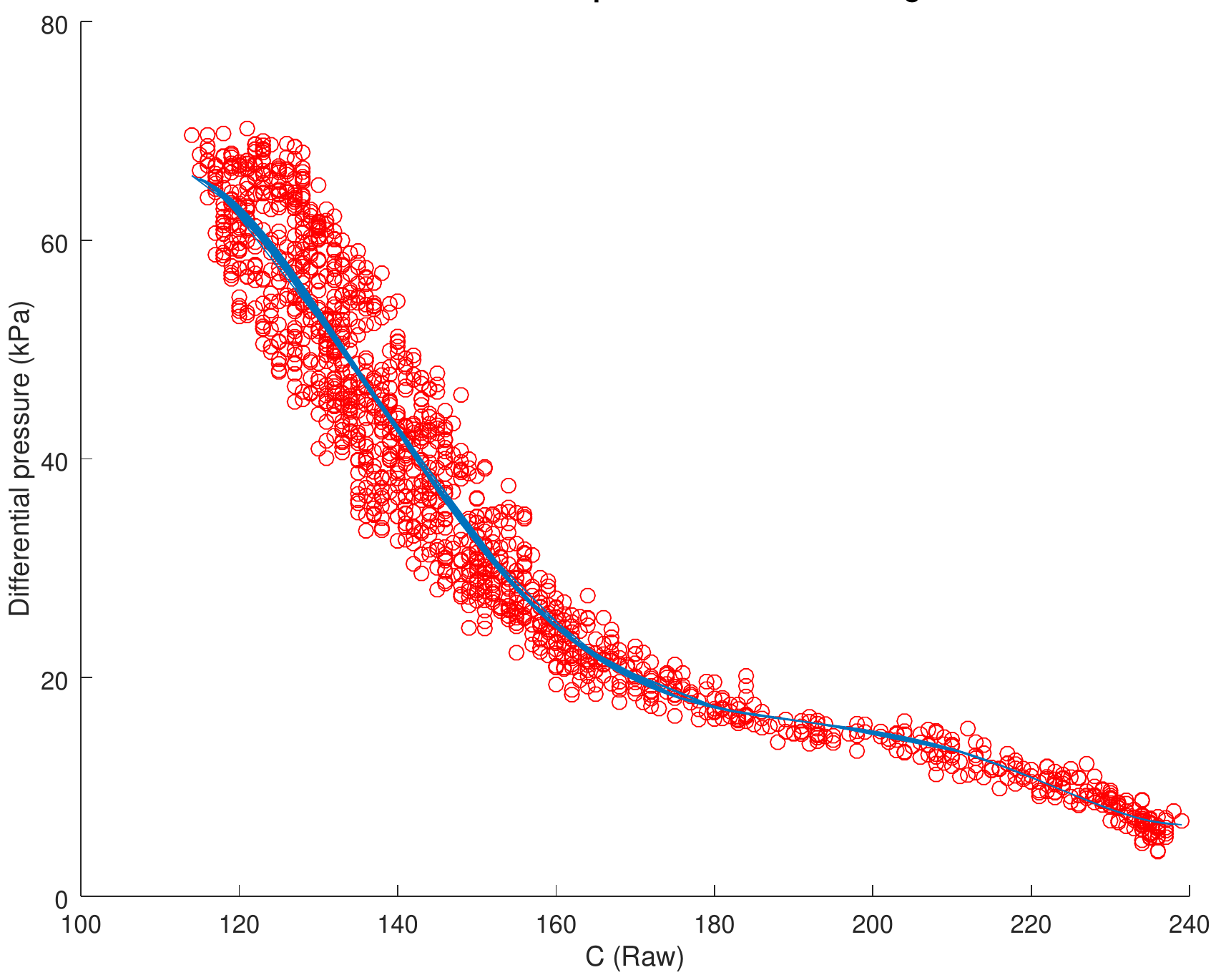}
\caption{Fifth order polynomial model (blue) fit to data points (red) in order to relate capacitance to pressure for an individual sensor. The fifth order polynomial model is found for each sensor during the calibration.}
\label{fig:poly_fit}
\end{figure}

Figure \ref{fig:poly_fit} depicts how the fifth order polynomial model, indicated with blue, is fit to the data points, given as red circles. It was observed that all the sensors have slightly different responses, varying in noise level, gain, initial offset and even the shape of the curve.

\subsection{Validation}

\begin{figure}[t!]
\centering
\includegraphics[width=0.35\textwidth]{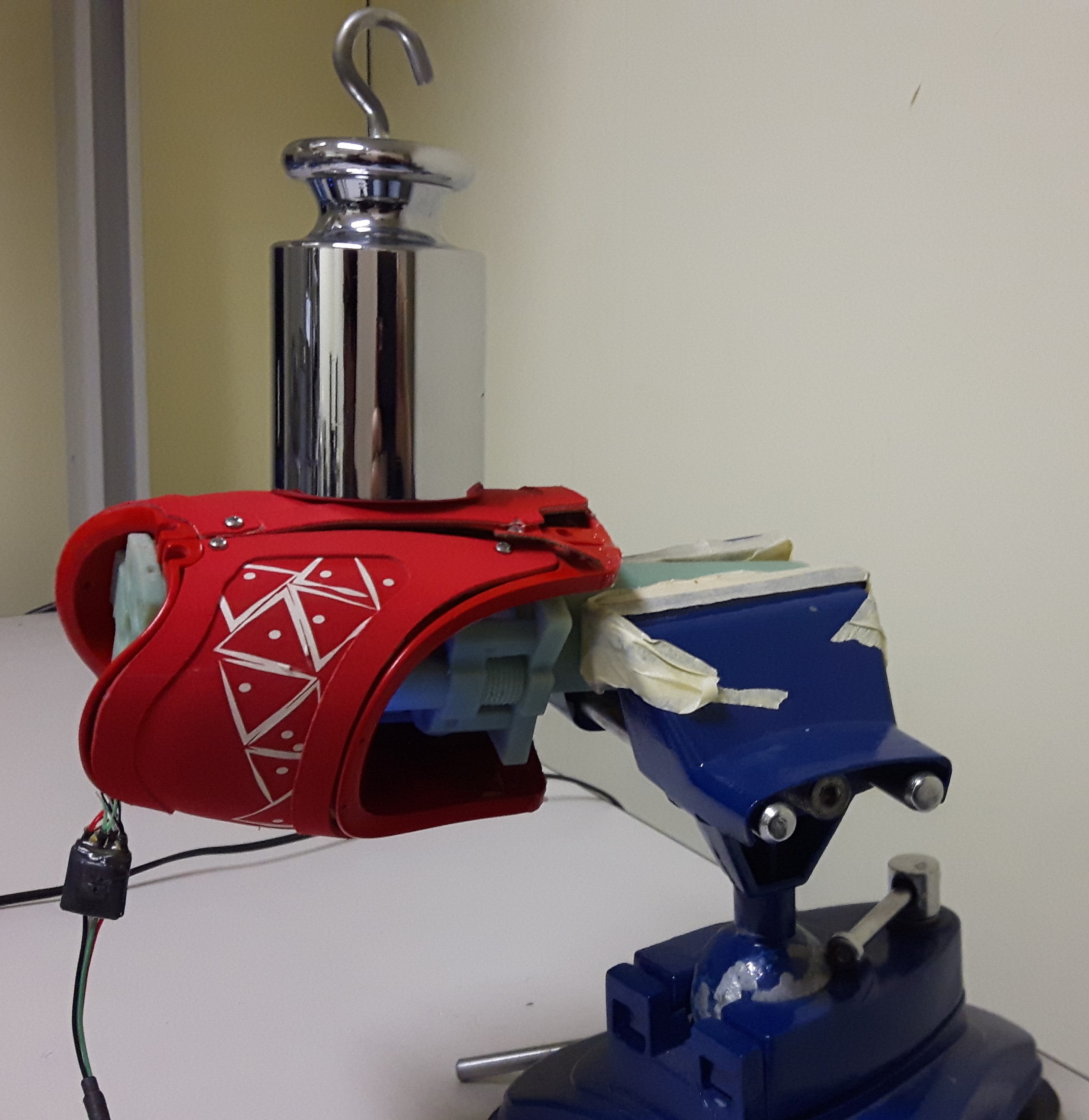}
\caption{Validation setup. Different masses with known weights were placed on the skin. A level was used to ensure the weight vector is normal to the ground.}
\label{fig:validation_setup}
\end{figure}

\begin{figure}[t!]
\centering
\includegraphics[width=0.45\textwidth]{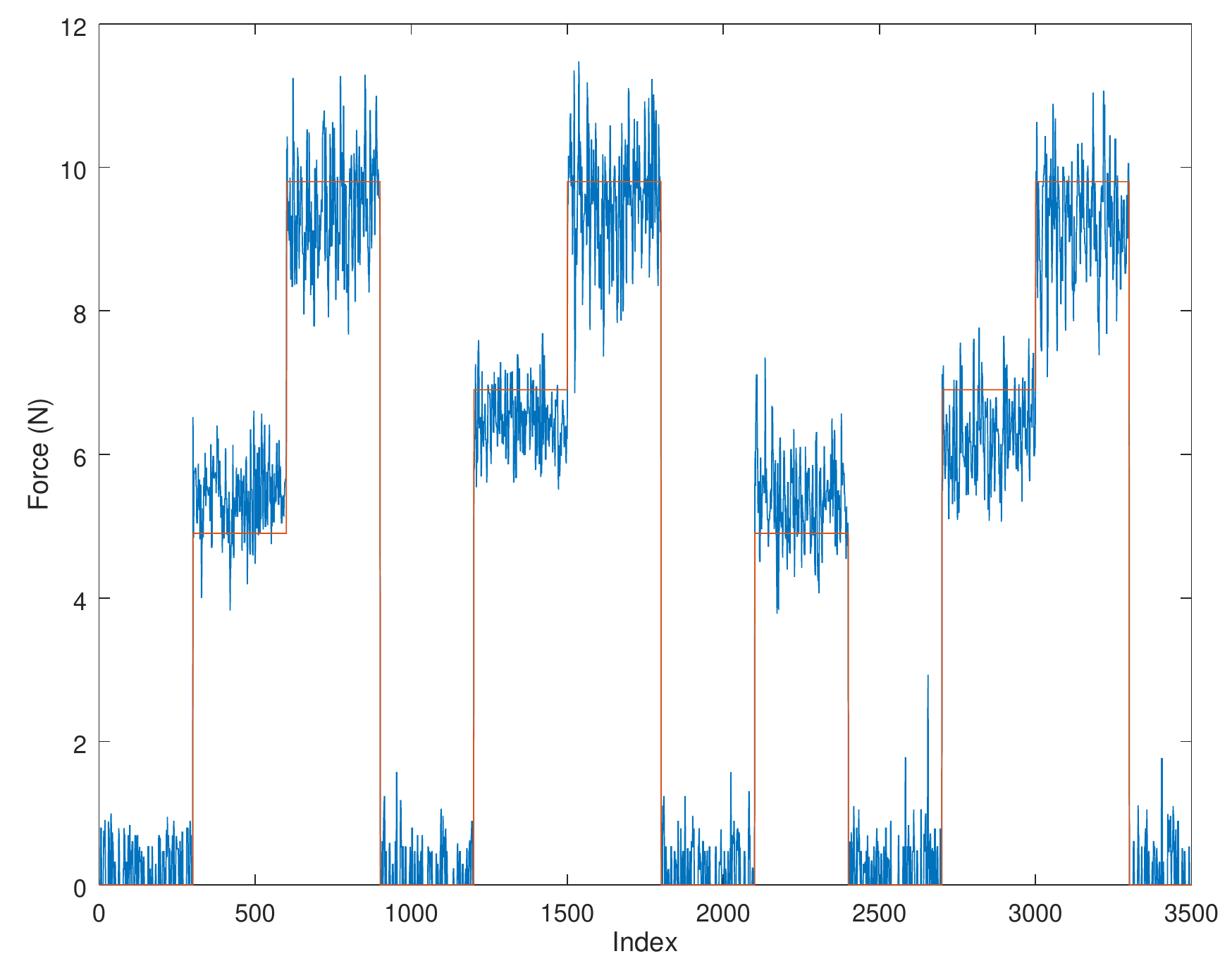}
\caption{Real force applied (red) compared with the estimated force by the sensors (blue). The mean relative error was around 13.2 \%.}
\label{fig:validation}
\end{figure}

The validation of the calibration results was performed by placing different masses with known weights on the skin, as shown in figure \ref{fig:validation_setup}, and comparing the applied force against the one estimated by the sensors. Using the fifth-order polynomial model for each individual sensor (Equation \ref{eq:polynomial}) to estimate the pressure applied to each one and using trilinear interpolation (Equation \ref{eq:totalForce}), that estimates the pressures in between the sensors, the total contact force applied to the skin can be calculated. The comparison between the ground truth and estimation can be seen in Figure \ref{fig:validation}. The graph indicates that the estimated force is close to the actual applied force (the mean relative error was around 13.2 \%) but the level of noise is high. This can be improved by using advanced filters.


\section{DISCUSSION AND CONCLUSION} \label{discussion_section}

This paper described a novel device that allows for the easy calibration of arrays of tactile sensors. The experiments were performed using the skin of iCub but other types of skins can be calibrated using the proposed device, after minor modifications are made to the software package. For example, other capacitive \cite{lee2006flexible}, piezoresistive \cite{kim2009polymer} and piezoelectric \cite{seminara2013piezoelectric} tactile sensor arrays could be calibrated using the device. In addition, the commercial packages \cite{commercial1,commercial2} can be re-calibrated if the force estimation has degraded over time.

The device can be used to accurately calibrate hundreds of sensors in just a couple of minutes with a small effort. The data indicated that the sensors do not saturate prematurely, as was the case for the vacuum bags experiment. The calibration results were validated using different masses that were placed on top of the skin and indicate acceptable errors in force estimation. Using the positive pressure to calibrate the skin does not restrict the maximum calibration range to be equal to 100 kPa (i.e. the atmospheric pressure), which is the case for the vacuum bag experiment that uses the negative pressure for calibration. The skins of various shapes and sizes can be calibrated using this device. 

However, there are multiple improvements that can be made to the current prototype of the calibration device in order to improve the performance. The following is the list of things that can be implemented in the near future:

\begin{enumerate}
    \item \textbf{Automatic activation of the the compressor:} a transistor, controlled by the microcontroller, can be added to the current circuitry in order to turn on the pump when the experiment starts and to switch it off once the calibration has finished.
    \item \textbf{More scalable software package:} in order to calibrate different types of skins with the device, multiple communication protocols and other configurations (e.g. different available mathematical models for the sensors, various filters to reject noise etc.) should be added to the current software package. 
    \item \textbf{Easier way to attach the lid:} currently the lid is attached using 8 screws which is a time consuming procedure. However, using some kind of clamping mechanism would make it faster to attach the lid to the chamber.
    \item \textbf{More compact electronics:} currently the electronics is on the breadboard that is messy, not robust and takes a lot of space. Converting the circuitry to a PCB would solve this issue. 
    \item \textbf{Better pressure sensor:} the current pressure sensor provides relatively noisy reading which degrades the calibration results. Using a pressure sensor that has better signal to noise ratio would enhance the device.
    \item \textbf{Better material for the chamber:} currently the material used for the chamber is 3D printed ABS and the air slightly leaks from the small holes within the structure. Using a material that can properly isolate air would solve this issue. 
    \item \textbf{Other material for the bladder:} the bladder used with the device is made out of low density polyethylene. It blocks the air from escaping but often breaks while installing or removing it from the device. Using a more durable material that can also block air flow would be a feasible solution.
    \item \textbf{Model for estimation error:} using the data logged during the calibration, it is possible to mathematically model the estimation error for each sensor at various pressures.
\end{enumerate}

Not all of those features are necessary but implementing these improvements would provide more robustness, easier set up and additional possibilities for the device.

\printbibliography

\end{document}